\documentclass[10pt,twocolumn,letterpaper]{article}

\usepackage{iccv}
\usepackage{times}
\usepackage{epsfig}
\usepackage{graphicx}
\usepackage{amsmath}
\usepackage{amssymb}
\usepackage{multirow}
\usepackage{float}
\usepackage{subfigure}

\usepackage[pagebackref=true,breaklinks=true,letterpaper=true,colorlinks,bookmarks=false]{hyperref}

\iccvfinalcopy 

\def\iccvPaperID{5} 

\ificcvfinal\fi

\begin{document}

\title{Simple Baseline for Single Human Motion Forecasting}

\author{Chenxi Wang$^\text{\dag}$, Yunfeng Wang, Zixuan Huang, Zhiwen Chen\\
Alibaba Group\\
{\tt\small \{lishui.wcx, weishan.wyf, zixuan.huangzixuan, zhiwen.czw\}@alibaba-inc.com}
}

\maketitle
\ificcvfinal\thispagestyle{empty}\fi

\begin{abstract}
   Global human motion forecasting is important in many fields, which is the combination of global human trajectory prediction and local human pose prediction. Visual and social information are often used to boost model performance, however, they may consume too much computational resources. In this paper, we establish a simple but effective baseline for single human motion forecasting \textbf{without visual and social information}, equipped with useful training tricks. Our method ``\textbf{futuremotion\_ICCV21}'' outperforms existing methods by a large margin on SoMoF benchmark\footnote{\url{https://somof.stanford.edu/results}\\\indent \ This work was done when Chenxi Wang was an intern at Alibaba Group.}. We hope our work provide new ideas for future research.
\end{abstract}


\section{Introduction}
Forecasting future human motion plays an important part in activity understanding, which is fundamental to many applications, including autonomous driving, human-computer interaction and public security. Previous work decomposes this problem to global human trajectory prediction~\cite{alahi2016social, gupta2018social, huang2019stgat} and local pose sequence prediction~\cite{mao2020history, mao2019learning, martinez2017human}. Recently, Adeli~\etal~\cite{adeli2020socially, adeli2021tripod} analyze the importance of jointly forecasting human trajectories and pose dynamics, and establish a new benchmark for further research. They also discuss the necessity of learning with visual and social information (i.e. human-human or human-object interactions).

Although image features and social interactions help human motion prediction, they may consume too much computational resources and pose a great challenge to model design. Instead, we try to improve the performance with simple pose sequence on a single person as input. Based on Graph Convolutional Networks (GCNs)~\cite{kipf2016semi}, we establish a simple but effective baseline for single human motion forecasting in the global scene. Many tricks are employed for effective training. Our method, named as ``\textbf{futuremotion\_ICCV21}'', outperforms existing methods by a large margin on SoMoF benchmark.
\section{Problem Definition}

Let $\mathbf{x}_t=(x_t^1, x_t^2, \cdots, x_t^J)$ be the global joint positions of one person at time $t$, where $x_t^i (i=1,2,\cdots,J)$ denotes the $i$-th joint coordinate of the body. Given $\mathbf{X}_{1:T} = (\mathbf{x}_1, \mathbf{x}_2, \cdots, \mathbf{x}_T)$, i.e., a whole series of single human pose from the 1st frame to the $T$-th frame, our goal is to forecast $\mathbf{X}_{T+1:T+\tau} = (\mathbf{x}_{T+1}, \mathbf{x}_{T+2}, \cdots, \mathbf{x}_{T+\tau})$, the global pose trajectories of the future $\tau$ frames.

``Global'' human pose trajectories indicate that the poses at different time share the same coordinate system. Joint positions can be in both 2D and 3D representations, where 2D joints are in pixel coordinate system and 3D joints are in world coordinate system.

Note that in our problem, images of input or output frames are \textbf{not} used for motion prediction. The prediction can be conducted only on a \textbf{single} person, which means we do not know other persons' motion in the same scene.
\section{Method}
In this section, we will introduce our main method and the tricks used to improve model performance, which contains backbone modification, data processing and training strategies.

\subsection{Backbone}

Following LTD~\cite{mao2019learning}, our method utilizes GCNs~\cite{kipf2016semi} combined with Discrete Cosine Transform (DCT), where the trajectories of a joint across all frames is encoded into frequency domain. Different from RNN-based method, the trajectories at all future frames are predicted in one time. The task is transferred to trajectory completion of $(\mathbf{X}_{1:T}||\mathbf{\hat{X}}_{T+1:T+\tau})$, where operator $||$ denotes concatenation, $\mathbf{\hat{X}}_{T+1:T+\tau}$ is the trajectory to be completed and padded using $\mathbf{x}_T$ in the input.

For $(x_1^j, x_2^j, \cdots, x_T^j)$, the $l$-th DCT coefficient $C_l^j$ is computed by
\begin{equation}
    \mathit{C_l^j} = \sqrt{\frac{2}{T}} \sum_{t=1}^{T} \frac{x_t^j}{\sqrt{1+\delta_{l1}}} \cos{\frac{\pi}{2T} (2t-1)(l-1)},
\end{equation}
where $l=1,2,\cdots,T$ and
\begin{equation}
    \delta_{ij} =
    \begin{cases}
        1 & \text{if } i = j,\\
        0 & \text{if } i\ne j.
    \end{cases}
\end{equation}
The sequence $(C_1^j, C_2^j, \cdots, C_T^j)$ is then considered as the input feature vector of joint $j$. GCN takes the features of all joints as input and output the frequency vector of the completed trajectory $(\tilde{C}_1^j, \tilde{C}_2^j, \cdots, \tilde{C}_T^j)$. Finally, the future trajectory is recovered from frequencies using inverse DCT:
\begin{equation}
    \tilde{x}_t^j = \sqrt{\frac{2}{T}} \sum_{l=1}^{T} \frac{\tilde{C}_l^j}{\sqrt{1+\delta_{l1}}} \cos{\frac{\pi}{2T} (2t-1)(l-1)}.
\end{equation}

Our experiments are conducted on LTD with several modifications in architecture and input representation.

\paragraph{XYZ as one node.}
\begin{figure}
    \centering
    \subfigure[XYZ as three nodes]
    {
    \centering
    \includegraphics[width=0.46\linewidth]{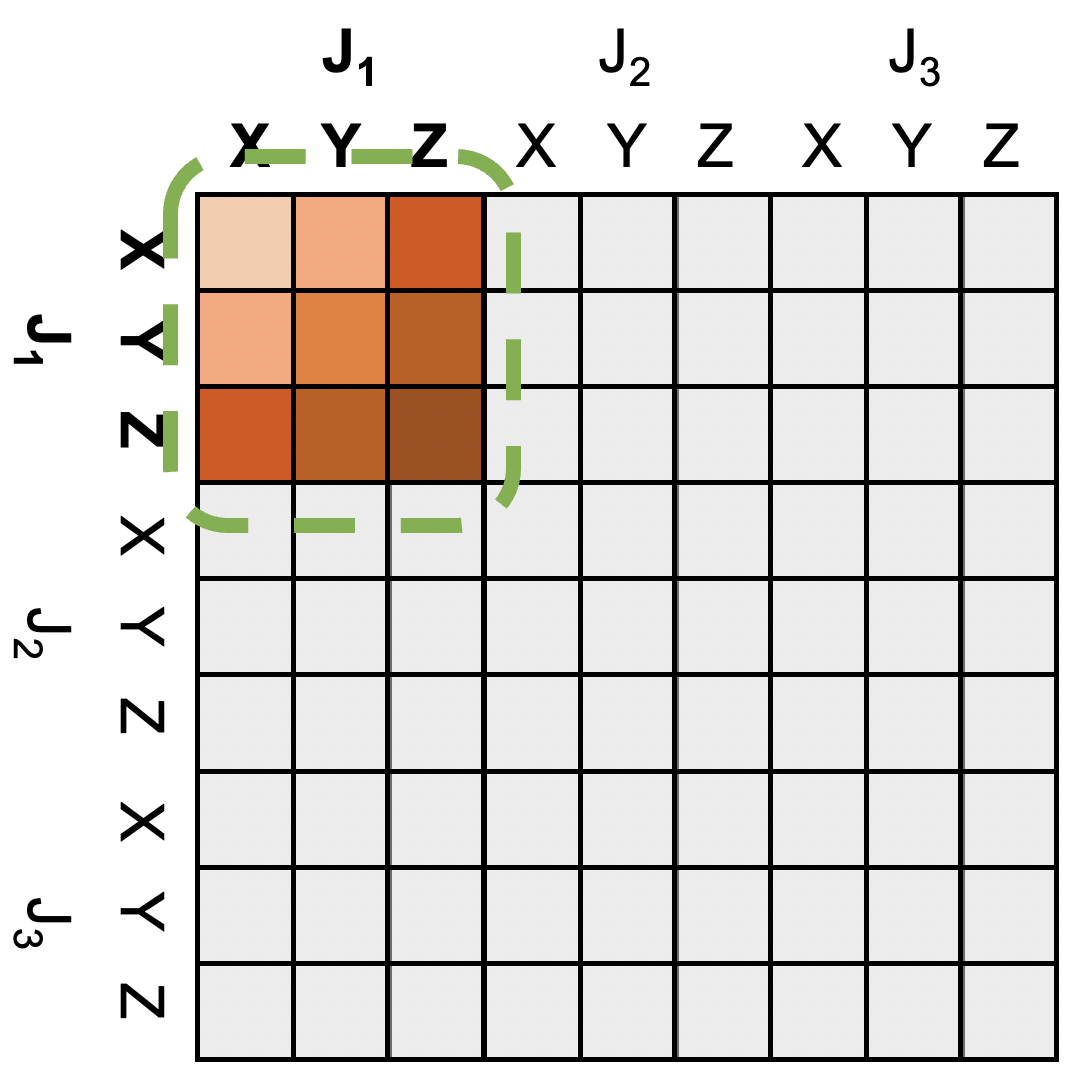}
    }
    \subfigure[XYZ as one node]
    {
    \includegraphics[width=0.46\linewidth]{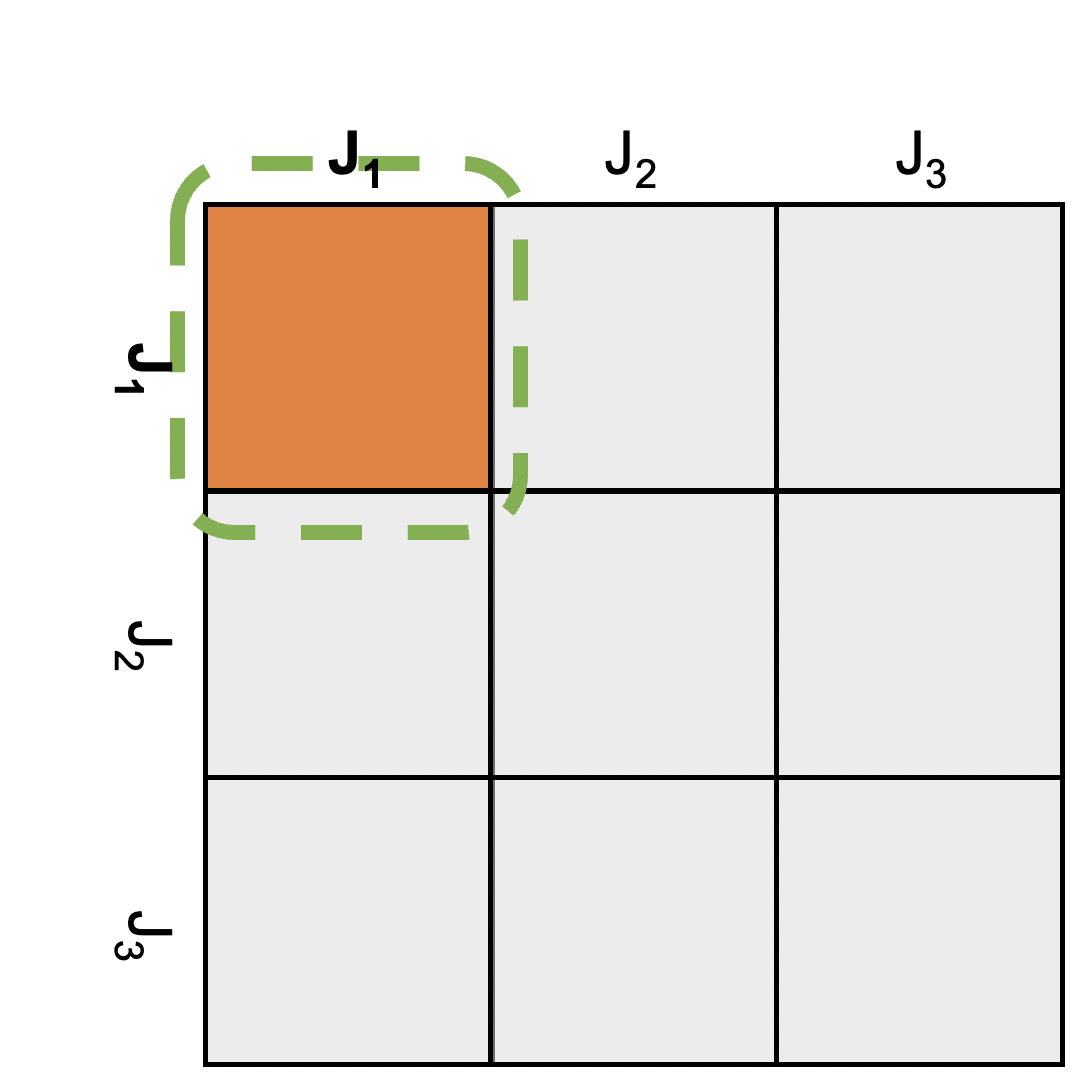}
    }
    \caption{Different node representations in adjacency matrix of GCN. A 3-joint graph has $9\times 9=81$ edges in the left matrix, but has only $3\times 3=9$ edges in the right matrix.}
    \label{fig:xyznode}
\end{figure}
In the original LTD implementation, different channels of a joint are treated as individual nodes in the graph, which means there are $J\times D$ nodes in a graph, where $D=2\text{ or }3$ is the dimensionality of the joint. Such representation ignores the association across different channels and hinders model performance. We instead bound different channels to the same node, and make them share the same edge weights. In Fig.~\ref{fig:xyznode}, there is an obvious reduction in the number of edges of adjacency matrix, which makes the edge weights easier to converge than before.

\paragraph{Position vs velocity.}
In LTD, frequency vector is generated from the joint positions directly, while velocities (position difference between neighbour frames) can also be used as input features. We make a comparison between position representation and velocity representation, and decide to use position representation for 2D task and velocity representation for 3D task.

\subsection{Data Processing}

\paragraph{Coordinate transform.}
Human pose trajectories are annotated in global coordinate systems (pixel coordinate system for 2D dataset and world coordinate system for 3D dataset). However, centralizing human positions can reduce coordinate range and boost model performance. Let $x_1^{j_c}$ be the human center at the 1st frame, the global human motions are centered by
\begin{equation}
    x_t^{j(c)} = x_t^j - x_1^{j_c} \ (t=1\text{ to }T, j=1\text{ to }J),
\end{equation}
where $j_c$ denotes the neck joint and $J$ the number of joints in human pose. Besides, we also scale the joint coordinates for easier convergence.

\paragraph{Visibility padding.}
Joint visibility is essential to the understanding of human-human relationship and human-object interaction in 2D motion prediction, since the occluded parts are caused by others from the environment. Intuitively, visibility can be predicted with the help of images, which is not allowed in our problem setting. Instead of predicting joint visibility using multi-modal input, we make a strong assumption: the visibility of a joint from frame $T+1$ to $T+\tau$ is the same as the one at frame $T$ (the last frame of input sequence). Based on this, we directly pad the visibility with the $T$-th frame. Let $v_t^j\in\{0,1\}$ denote the visibility of the $j$-th joint at the $t$-th frame, we generate the future visibility $\hat{v}_t^j$ by
\begin{equation}
    \hat{v}_t^j = v_T^j \ (t=T+1\text{ to }T+\tau, j=1\text{ to }J).
\end{equation}
While padding strategy is simple to implement, it is effective and accurate for most scenes, which establishes a strong baseline for future research.

\paragraph{Interpolation of invisible joints.}
\begin{figure}
    \centering
    \includegraphics[width=\linewidth]{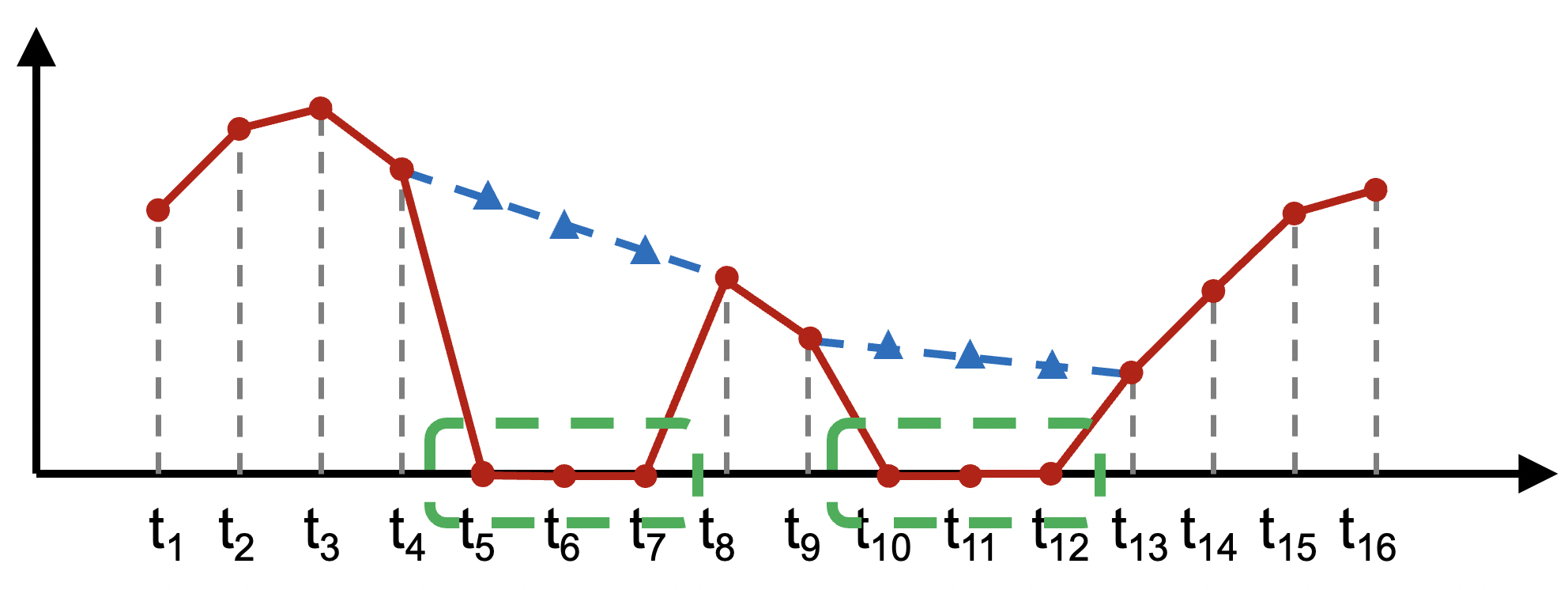}
    \caption{Interpolation of invisible joints. The red line shows the trajectory of a joint with a few missing frames (in green box). The blue lines shows the results of binary interpolation.}
    \label{fig:interp}
\end{figure}
Some joint coordinates are padded with zeros when they are invisible in an image, which keeps the consistency of data format. However, the incoherence in pose trajectories brings more instability and makes the model hard to converge. To avoid missing data in pose trajectories, we adopt linear interpolation in both training and inference (Fig.~\ref{fig:interp}). For a pose sequence $(x_t^j, x_{t+1}^j, \cdots, x_{t+m}^j)$ with $x_{t+i}^j=0\ (i=1\text{ to }m-1)$, coordinates are interpolated by
\begin{equation}
    x_{t+i}^{j*} = \frac{i}{m}\times (x_{t+m}^j - x_t^j) + x_t^j,
\end{equation}
where $x_{t+i}^{j*}$ stands for the interpolated value of joint $j$ at frame $t+i$. Although linear interpolation may not be suitable for all trajectories, there is still an obvious decrease in prediction errors.

\paragraph{Boundary filtering.}
\begin{figure}
    \centering
    \subfigure[$t_1$]
    {
    \centering
    \includegraphics[width=0.46\linewidth]{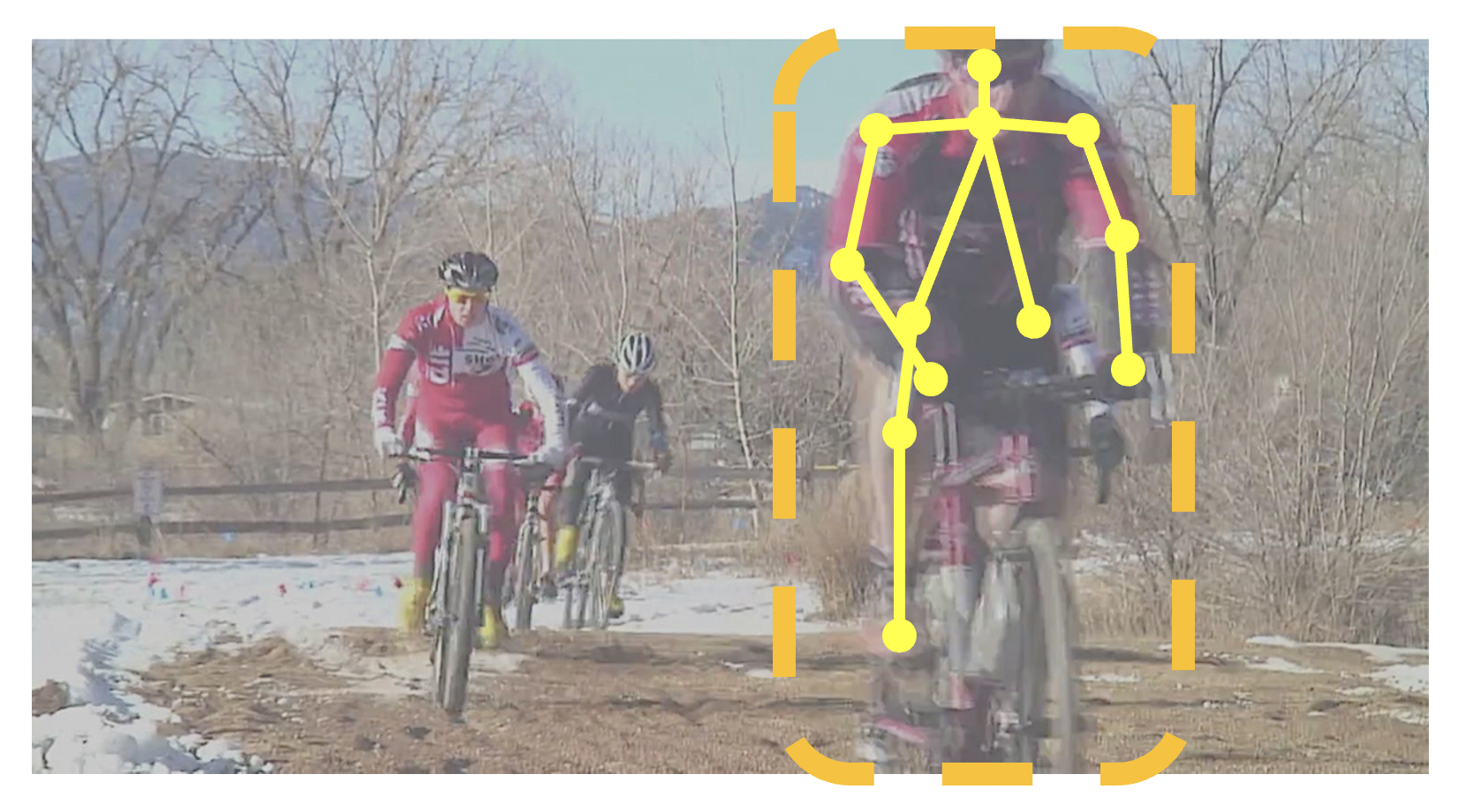}
    }
    \subfigure[$t_2$]
    {
    \includegraphics[width=0.46\linewidth]{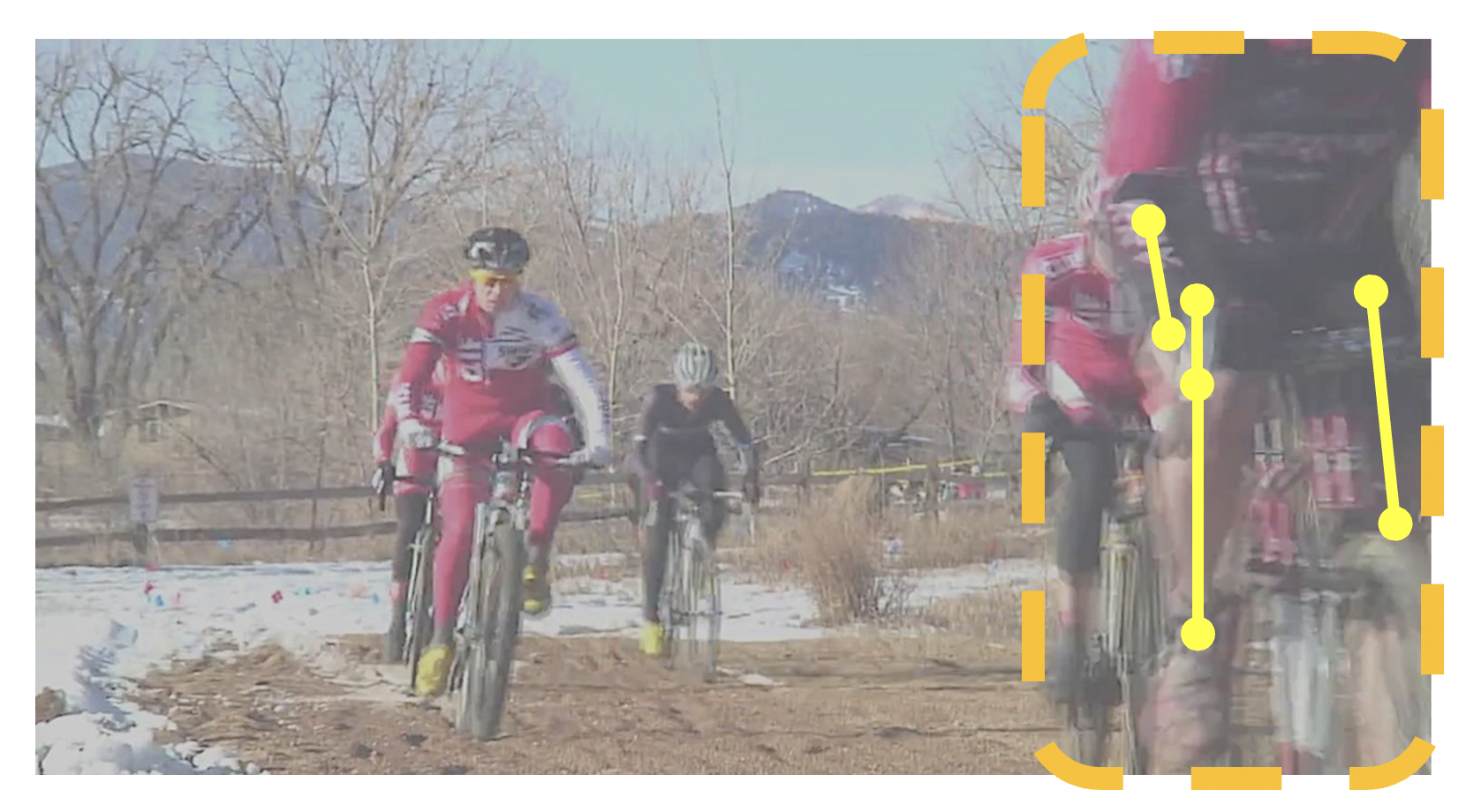}
    }
    \caption{Joints truncated by image boundary. Compared with frame $t_1$, the person in the orange box of frame $t_2$ loses half of the joints because he goes out of the camera sight.}
    \label{fig:boundary}
\end{figure}
In some 2D scenes, joints become invisible when a person is out of camera sight (Fig.~\ref{fig:boundary}). Such cases cannot be detected by our models, and brings unavoidable errors in final inference. Since such cases happen with an obvious trend, the joint positions output by our model often cross the image boundary correspondingly. Based on this, we adopt a boundary filtering strategy in post-processing, where the visibility of all the predicted 2D joints out of camera sight are set to 0.

\subsection{Training}

\begin{figure}
    \centering
    \includegraphics[width=\linewidth]{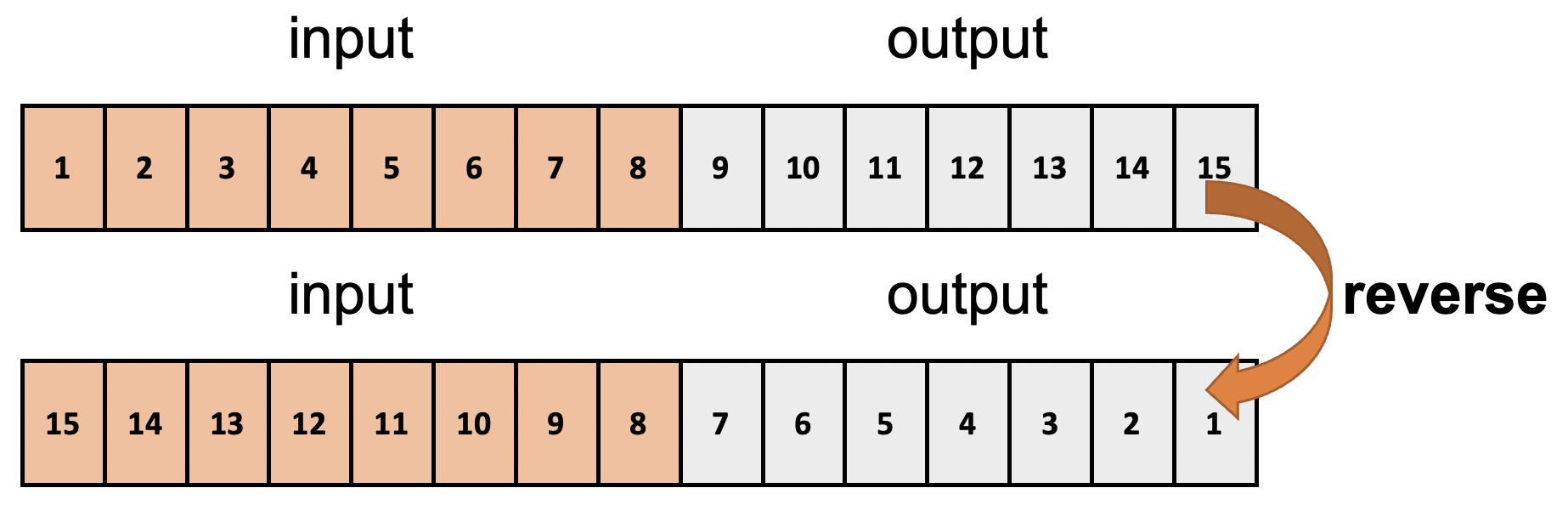}
    \caption{An example of sequence reversing with 8-frame input and 7-frame output.}
    \label{fig:dataaug}
\end{figure}

\paragraph{Data augmentation.}
Since human motion is a sequence of body joints, the reversed sequence can also be used in model training. The input sequence $\mathbf{X}_{1:T}$ and label sequence $\mathbf{X}_{T+1:T+\tau}$ are concatenated first, followed by a reverse operation. Frame $1$ to $T$ in the reversed sequence are taken as the new input and other frames as the new label. Fig.~\ref{fig:dataaug} is an example of $T=8$ input frames (data) and $\tau=7$ output frames (label). During model training of 3D motion forecasting, we randomly flip the pose sequence to generate new data. 
\begin{table*}[]
    \centering
    \scalebox{0.8}{
    \begin{tabular}{c||c|ccccc||c|ccccc}
        \hline
        \multirow{3}{*}{Method} & \multicolumn{6}{c||}{3DPW (VIM)} & \multicolumn{6}{c}{PoseTrack (VAM)}  \\
        \cline{2-13}
         & \multirow{2}{*}{Average} & \multicolumn{5}{c||}{milliseconds} & \multirow{2}{*}{Average} & \multicolumn{5}{c}{milliseconds} \\
         & & 100 & 240 & 500 & 640 & 900 & & 80 & 160 & 320 & 400 & 560 \\
         \hline
         PF-RNN~\cite{martinez2017human} + S-LSTM~\cite{alahi2016social} & 172.09 & 73.82 & 127.23 & 179.07 & 202.78 & 277.55 & 127.95 & 89.44 & 111.11 & 136.43 & 145.72 & 157.04 \\
         PF-RNN~\cite{martinez2017human} + S-GAN~\cite{gupta2018social} & 180.29 & 83.35 & 138.48 & 182.84 & 204.84 & 291.96 & 122.99 & 87.23 & 106.23 & 131.12 & 139.94 & 150.44 \\
         PF-RNN~\cite{martinez2017human} + ST-GAT~\cite{huang2019stgat} & 158.69 & 66.95 & 117.77 & 165.99 & 190.52 & 252.23 & 121.18 & 86.76 & 102.61 & 127.87 & 137.87 & 150.80 \\
         Mo-Att~\cite{mao2020history} + S-LSTM~\cite{alahi2016social} & 162.96 & 64.64 & 111.67 & 168.67 & 202.16 & 267.65 & 124.83 & 86.76 & 109.02 & 130.82 & 142.35 & 155.21 \\
         Mo-Att~\cite{mao2020history} + S-GAN~\cite{gupta2018social} & 166.48 & 66.36 & 112.18 & 166.48 & 209.53 & 277.85 & 121.89 & 85.82 & 104.13 & 128.97 & 139.07 & 151.45 \\
         Mo-Att~\cite{mao2020history} + ST-GAT~\cite{huang2019stgat} & 150.21 & 62.15 & 97.74 & 155.23 & 184.96 & 250.98 & 119.59 & 86.29 & 100.92 & 125.31 & 137.50 & 147.92 \\
         \hline
         SC-MPF~\cite{adeli2020socially} & 123.94 & 46.28 & 73.89 & 130.24 & 160.84 & 208.45 & 117.80 & 78.36 & 99.80 & 124.38 & 138.52 & 147.93 \\
         TriPOD~\cite{adeli2021tripod} & 83.66 & 30.27 & 51.84 & 85.09 & 104.79 & 146.33 & 72.74 & 30.00 & 49.66 & 80.32 & 93.32 & 110.40 \\
         \hline
         Ours & \textbf{49.40} & \textbf{9.49} & \textbf{22.89} & \textbf{50.94} & \textbf{66.22} & \textbf{97.44} & \textbf{51.76} & \textbf{19.03} & \textbf{32.53} & \textbf{56.23} & \textbf{67.83} & \textbf{83.17}\\
         \hline
    \end{tabular}
    }
    \caption{Evaluation results on 3DPW and PoseTrack testing sets.}
    \label{tab:main_results}
\end{table*}
\begin{table*}[]
    \centering
    \scalebox{0.8}{
    \begin{tabular}{c||c|ccccc||c|ccccc}
        \hline
        \multirow{3}{*}{Optimization} & \multicolumn{6}{c||}{3DPW (VIM)} & \multicolumn{6}{c}{PoseTrack (VAM)} \\
        \cline{2-13}
         & \multirow{2}{*}{Average} & \multicolumn{5}{c||}{milliseconds} & \multirow{2}{*}{Average} & \multicolumn{5}{c}{milliseconds} \\
         & & 100 & 240 & 500 & 640 & 900 & & 80 & 160 & 320 & 400 & 560 \\
         \hline
         LTD & 69.47 & 20.05 & 40.53 & 78.17 & 91.30 & 117.31 & 100.01 & 70.18 & 83.92 & 105.09 & 113.80 & 127.07  \\
         Coordinate transform & 52.68 & 15.88 & 31.50 & 57.81 & 68.81 & 89.43 & 99.41 & 69.83 & 83.40 & 104.16 & 112.80 & 126.87 \\
         Visibility Padding & --- & --- & --- & --- & --- & --- & 62.11 & 22.96 & 41.48 & 69.18 & 80.38 & 96.57 \\
         XYZ as one node & 51.21 & 15.40 & 30.41 & 56.03 & 66.88 & 87.35 & 61.98 & 22.41 & 40.94 & 69.07 & 80.59 & 96.91 \\
         Curriculum learning & 48.50 & 14.32 & 28.14 & 52.93 & 63.61 & 83.49 & 60.58 & 21.42 & 39.52 & 67.18 & 78.97 & 95.82 \\
         Velocity input & 47.36 & 13.60 & 27.53 & 51.91 & 62.20 & 81.54 & --- & --- & --- & --- & --- & --- \\
         Interpolation of invisible joints & --- & --- & --- & --- & --- & --- & 59.37 & 20.88 & 38.27 & 65.55 & 77.36 & 94.80 \\
         Data augmentation & 46.48 & 13.53 & 27.03 & 50.65 & 60.91 & 80.28 & --- & --- & --- & --- & --- & --- \\
         Data extension & 38.86 & 7.01 & 18.18 & 42.59 & 53.34 & 73.16 & 57.18 & 19.99 & 36.78 & 63.15 & 74.59 & 91.41 \\
         Boundary filtering & --- & --- & --- & --- & --- & --- & 55.26 & 20.10 & 35.80 & 60.77 & 71.62 & 87.99 \\
         OHKM loss & 38.78 & 6.93 & 18.17 & 42.54 & \textbf{53.29} & \textbf{72.99} & \textbf{55.20} & 19.88 & 36.58 & \textbf{60.69} & \textbf{71.27} & \textbf{87.59} \\
         Short-term optimization & \textbf{38.66} & \textbf{6.57} & \textbf{17.65} & \textbf{42.40} & 53.35 & 73.33 & 60.39 & \textbf{19.54} & \textbf{36.11} & 69.70 & 80.83 & 95.78 \\
         \hline
         Fusion & \textbf{38.58} & \textbf{6.57} & \textbf{17.65} & \textbf{42.40} & \textbf{53.29} & \textbf{72.99} & \textbf{55.04} & \textbf{19.54} & \textbf{36.11} & \textbf{60.69} & \textbf{71.27} & \textbf{87.59}\\
         \hline
    \end{tabular}
    }
    \caption{Model improvements with cumulative optimizations on 3DPW and PoseTrack validation sets. ``---'' indicates skipping this step.}
    \label{tab:optim}
\end{table*}

\begin{figure}
    \centering
    \includegraphics[width=\linewidth]{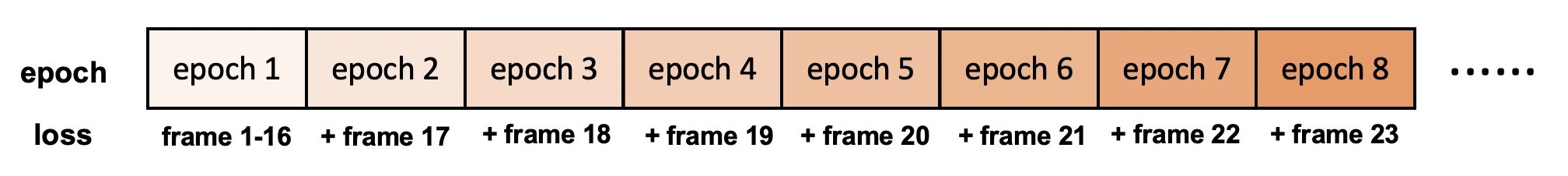}
    \caption{Curriculum learning for motion forecasting. The frames are added to loss computation step by step.}
    \label{fig:curriculum}
\end{figure}

\paragraph{Curriculum learning~\cite{bengio2009curriculum}.}
This approach is employed in many machine learning tasks, which is to learn model parameters from easy samples first, and add harder ones step by step. Since long-term human poses are more difficult to predict than short-term ones, curriculum learning can also be used in motion prediction~\cite{adeli2021tripod}. Here we adopt a similar strategy as \cite{adeli2021tripod}. As Fig.~\ref{fig:curriculum} shows, for a concatenated sequence $(\mathbf{X}_{1:T} || \mathbf{X}_{T+1:T+\tau})$, we only use frame $1$ to $T$ in loss computation in the first $E$ training steps, then add frame $T+t$ to the loss at the $(E\times t)$-th step, where $t=1\text{ to }\tau$.

\paragraph{OHKM loss.}
In training we firstly use default smooth-$l_1$ as the loss function, while we discover that the loss of some joints are harder to decrease than other joints. Hence, we adopt online hard keypoints mining (OHKM) approach~\cite{chen2018cascaded} in loss computation, which is to dynamically consider the joints with top-$k$ losses in training and set the loss of other joints to 0. This strategy helps the model to concentrate more on hard samples and effectively reduce prediction errors. We use top-8 joints in PoseTrack dataset and top-6 joints in 3DPW dataset.

\vspace{-0.05in}
\paragraph{Short-term motion optimization.}
As is proven in previous research~\cite{mao2020history, mao2019learning}, long-term pose trajectory prediction is harder to model than short-term counterpart. We also discover that training for long-term motion decreases the performance of short-term predictions, which indicates a trade-off between them when using only a single model. Instead, we adopt two models with the same architecture, to predict the short-term motion and long-term motion respectively, in which way short-term prediction will not be influenced. In our experiments, we take the first 4 frames as the short-term sequence, and make a result fusion in the final inference.

\vspace{-0.05in}
\paragraph{Data extension.}
The original pose sequences provided by the dataset are cut from coherent video frames, which inspires us to find a way to expand the data. Sequences from the same video are concatenated to be a long one. We randomly select a frame $t_0$ as the start frame and takes the pose trajectory $(\mathbf{X}_{t_0:t_0+T-1} || \mathbf{X}_{t_0+T:t_0+T+\tau-1})$ as one training sample, in which way data can be well extended. We also use AMASS~\cite{mahmood2019amass} dataset for model pretraining in 3D task. We notice that there are only a few annotated frames in PoseTrack dataset, so we adopt Halpe pretrained model in AlphaPose tracking module~\cite{fang2017rmpe, li2018crowdpose, li2020pastanet, xiu2018poseflow} to annotate the rest frames and use them for training in 2D task.
\section{Experiments}

\begin{table*}[]
    \centering
    \begin{tabular}{ccc|c||ccc|c}
         \hline
         \multicolumn{3}{c|}{Parameters} & 3DPW (VIM) & \multicolumn{3}{c|}{Parameters} & PoseTrack (VAM)\\
         \cline{1-3}\cline{5-7}
         {\ \ \ \ Scale\ \ \ } & {\ \ \#Block\ \ } & {\#Channel} & Average & {\ \ \ \ Scale\ \ \ } & {\ \ \#Block\ \ } & {\#Channel} & Average\\
         \hline
         10 & 12 & 256 & 39.24 & 0.5 & 12 & 256 & 55.30 \\
         50 & 12 & 256 & 38.90 & 1 & 12 & 256 & 55.20 \\
         100 & 12 & 256 & 38.78 & 5 & 12 & 256 & 55.29 \\
         \hline
         100 & 8 & 256 & 39.13 & 1 & 8 & 256 & 55.43 \\
         100 & 12 & 256 & 38.78 & 1 & 12 & 256 & 55.20 \\
         100 & 16 & 256 & 38.97 & 1 & 16 & 256 & 55.26\\
         \hline
         100 & 12 & 128 & 39.59 & 1 & 12 & 128 & 55.75\\
         100 & 12 & 256 & 38.78 & 1 & 12 & 256 & 55.20 \\
         100 & 12 & 512 & 38.82 & 1 & 12 & 512 & 55.34\\
         \hline
    \end{tabular}
    \caption{Evaluation of models with different parameters on 3DPW and PoseTrack validation sets.}
    \label{tab:params}
\end{table*}

In this section, we first detail our experimental settings, followed by the comparison between our method and other representative methods. Then we explore the model performance with different parameters. Finally, the optimization steps across the whole process are listed in detail.

\subsection{Experimental Settings}

\paragraph{Implementation details.} We use the same GCN architecture with LTD~\cite{mao2019learning}, with 30-d DCT matrix, 12 GCN blocks and 256-d hidden features. All the models are trained for 50 epochs with Adam~\cite{kingma2014adam} optimizer, where the initial learning rate is set to 0.001, and multiplied by 0.95 every one epoch. For curriculum learning, we add one frame to the loss every two epochs.

\paragraph{Datasets.} For 2D motion forecasting, we use PoseTrack~\cite{andriluka2018posetrack} for training and evaluation, which is widely used in research on pose estimation and pose tracking, and contains both pose sequences and joint visibility annotations. For 3D motion forecasting, we use 3DPW~\cite{vonMarcard2018} dataset, which contains a large amount of accurate 3D poses in 60 scenes. Besides, we also use AMASS~\cite{mahmood2019amass} dataset for pretraining in 3D task, which is a large database of various human motions. Following \cite{adeli2021tripod}, we use $T=16$ and $\tau=14$ in both 2D and 3D tasks.

\paragraph{Evaluation metrics.} Following \cite{adeli2021tripod}, we use Visibility-Ignored Metric (\textbf{VIM}) for 3D motion forecasting and Visibility-Aware Metric (\textbf{VAM}) for 2D motion forecasting in evaluation. VIM aims to compute the position error only on visible joints, and invisible joints are not penalized. Since 3DPW has no invisible joints, VIM becomes similar to MPJPE~\cite{ionescu2013human3} metric. VAM computes joint position error when visibility prediction is correct, otherwise a penalty item $\beta$ will be added to the error. In evaluation, VIM is scaled from meters to centimeters and $\beta=200$ in VAM calculation.

\subsection{Main Results}
\begin{figure*}
    \centering
    
    \includegraphics[width=\linewidth]{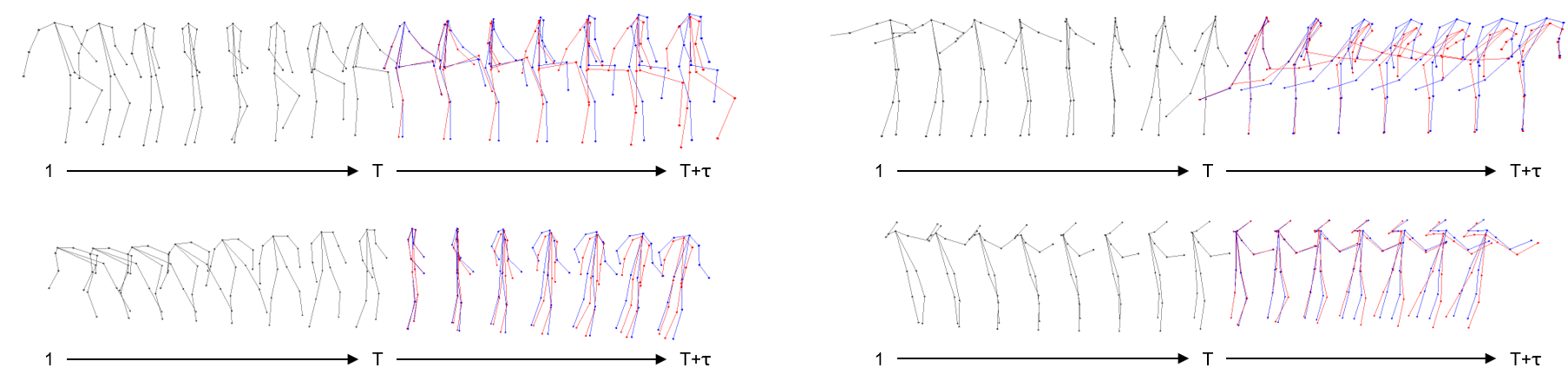}
    \caption{Visualization of results on 3DPW dataset. Poses are sampled every two frames from frame 1 to frame $T+\tau$. The gray poses indicate the input motion, and the blue ones and red ones stand for predictions and ground truth respectively.}
    \label{fig:results_3dpw}
    \vspace{0.1in}
    \includegraphics[width=\linewidth]{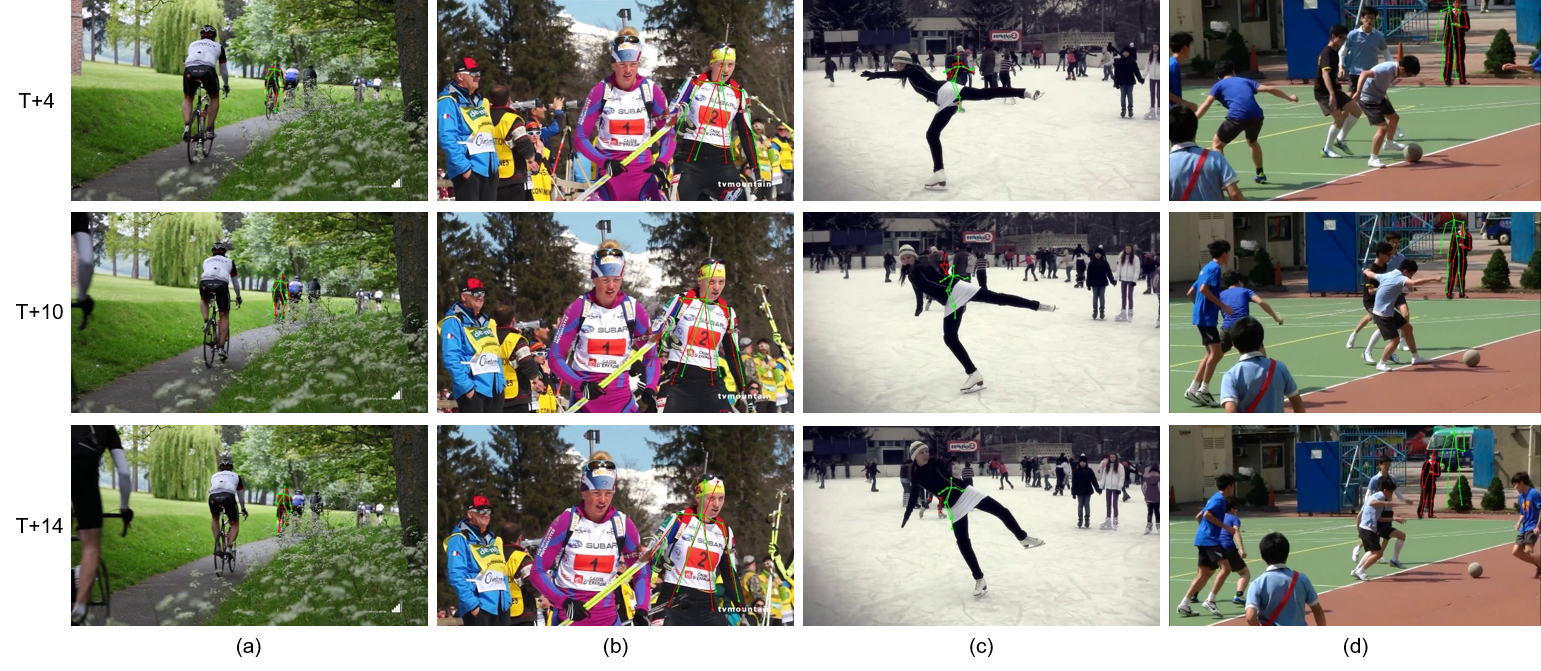}
    \caption{Visualization of results on PoseTrack dataset. We visualize the predictions (green) and ground truth (red) at frame 4, 10 and 14 of the output sequences. (c) and (d) show the results in scenes with occlusion and camera shake respectively. Note that the images are only used for clearer display, but not used in our training and inference.}
    \label{fig:results_posetrack}
\end{figure*}
Our method is compared with two lines of representative methods shown on SoMoF leaderboard. The first line is the combination of global human trajectory prediction (S-LSTM~\cite{alahi2016social}, S-GAN~\cite{gupta2018social} and ST-GAT~\cite{huang2019stgat}) and local pose dynamic prediction (PF-RNN~\cite{martinez2017human} and Mo-Att~\cite{mao2020history}), and the other line adopts direct prediction for global human pose sequence.

The results of these approaches on 3DPW and PoseTrack are detailed in Tab.~\ref{tab:main_results}, where VIM and VAM are computed at different frames. Our method outperforms all the previous ones across both short-term and long-term predictions by a large margin, which proves the effectiveness of this simple baseline model.

Fig.~\ref{fig:results_3dpw} and~\ref{fig:results_posetrack} visualize the results on 3DPW and PoseTrack respectively, which indicates long-term motions are usually harder to predict than short-term motions. We can also find that occlusion and camera shake make 2D motion forecasting more difficult.

\subsection{Step-by-step Optimization}
\vspace{-0.1in}
Tab.~\ref{tab:optim} records our attempts to improve model performance in the whole process. During exploration, we found that the improvements on PoseTrack are less obvious compared with 3DPW. On the one hand, 2D coordinates and 3D counterparts do not follow the same translation rules due to the difference of coordinate systems. On the other hand, the manual annotations in PoseTrack may be imprecise and decrease the effectiveness of these tricks.

\subsection{Model Parameters}
We change the coordinate scale, number of GCN blocks and number of hidden layer channels and make a comparison on 3DPW and PoseTrack validation sets. The results are demonstrated in Tab.~\ref{tab:params}. We choose the parameters with the minimum validation errors as the main training setting, and other models are used in the final result fusion.
\section{Conclusion}
We establish a simple but effective baseline for single human motion forecasting without visual and social information. Practical tricks are applied to reduce prediction error. Experiments demonstrate the effectiveness of our method and optimization strategies. Our method significantly surpasses existing methods on SoMoF benchmark. We hope our work provide more inspirations for future research on both single and social human motion forecasting.

{\small
\bibliographystyle{ieee_fullname}
\bibliography{egbib}
}

\end{document}